\documentclass[10pt,letterpaper]{article}

\usepackage[top=0.85in,left=2.75in,footskip=0.75in]{geometry}
\usepackage{amsmath,amssymb}
\usepackage{changepage}
\usepackage{textcomp,marvosym}
\usepackage{cite}
\usepackage{nameref,hyperref}
\usepackage[nopatch={eqnum,footnote}]{microtype}
\DisableLigatures[f]{encoding = *, family = * }
\usepackage[table]{xcolor}
\usepackage{array}
\usepackage{booktabs}

\newcolumntype{+}{!{\vrule width 2pt}}
\newlength\savedwidth

\usepackage{setspace}
\raggedright
\usepackage[
  aboveskip=1pt,
  labelfont=bf,
  labelsep=period,
  justification=raggedright,
  singlelinecheck=off
]{caption}

\makeatletter
\renewcommand{\@biblabel}[1]{\quad#1.}
\makeatother

\usepackage{lastpage,fancyhdr,graphicx}
\usepackage{epstopdf}
\fancyheadoffset[L]{2.25in}
\fancyfootoffset[L]{2.25in}
\begin{document}
\vspace*{0.2in}

\begin{flushleft}
{\Large
\textbf{Abstracts or full text? Evaluating input scope and computational efficiency in biomedical lay summarization}
}
\newline
\\
Egecan Çelik Evgin\textsuperscript{1\Yinyang*},
\.{I}lknur Karadeniz\textsuperscript{2\Yinyang},
Olcay Taner Yıldız\textsuperscript{1,3\Yinyang}
\\
\bigskip

\textbf{1} Department of Computer Science, Özyeğin University, İstanbul, Türkiye
\\
\textbf{2} Department of Computer Engineering, Galatasaray University, İstanbul, Türkiye
\\
\textbf{3} Department of Artificial Intelligence and Data Engineering, Özyeğin University, İstanbul, Türkiye
\\
\bigskip

\Yinyang All authors contributed equally to this work.
\\
* egecan.evgin@ozu.edu.tr; ikaradeniz@gsu.edu.tr; olcay.yildiz@ozyegin.edu.tr
\\

\end{flushleft}


\section*{Abstract}

Biomedical research articles are mainly written for experts and often contain terms that are difficult for the general public to understand. Automated lay summarization has become a solution to make biomedical knowledge easier for patients and the public to understand. However, generating lay summaries from full-text articles requires more computing power because the articles are long and costly to process. Fine-tuning large language models for this task also requires substantial resources. This study asks two questions: (1) whether abstracts contain enough information to produce lay summaries as good as those based on full-text articles, and (2) whether few-shot learning performs as well as task-specific fine-tuning. The study used two public biomedical lay summarization datasets and two recent models, evaluated for readability, factual consistency, semantic similarity, and runtime efficiency. Results show that abstract-based summarization performs better overall. Across both datasets and models, abstract fine-tuning is 4.70 times faster, while abstract zero- and few-shot prompting is 5.31 times faster than full-text methods. These results show that abstract-based prompting reduces runtime cost while maintaining lay summary quality.

\clearpage
\newgeometry{top=0.85in,left=1in,right=1in,footskip=0.75in}

\section*{Introduction}
Biomedical research articles are important means of communicating research findings in the life sciences \cite{forero2020brief}. These articles are highly structured and often use dense technical language that can be difficult for non-expert readers to understand \cite{sollaci2004imrad,devaraj2021paragraph}. Lay summaries have become a useful way to make biomedical research more accessible to these readers \cite{nunn2014lay,king2017plain}. Unlike standard summaries, lay summaries prioritise readability and use simpler language, reducing domain-specific terminology and enabling non-expert readers to understand the main findings \cite{rosenberg2021open,wen2025plain}. However, automatically generating lay summaries is often more challenging, as it requires models to improve readability while keeping the summaries relevant and factual \cite{alhussaini2023pathology,rosati2023grasum}. For example, Figure 1 illustrates the differences between technical and lay summaries of the same biomedical research article.

\begin{figure}[!ht]
    \centering
    \includegraphics[width=\textwidth]{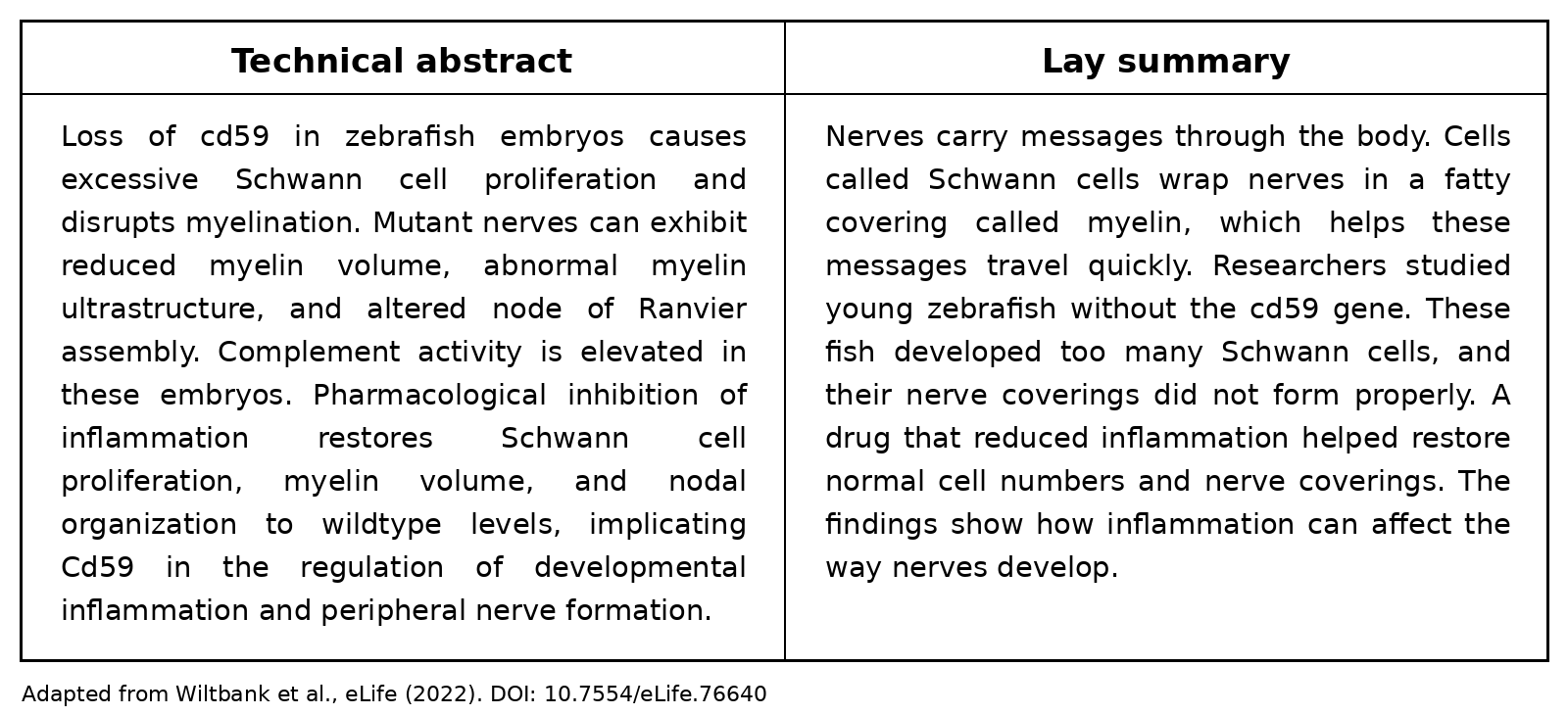}
    \caption{\textbf{Technical and lay descriptions of biomedical research.}
    Shortened versions adapted from the abstract and eLife digest
    of Wiltbank et al.~\cite{wiltbank2022cd59}.}
    \label{fig:technical-lay-example}
\end{figure}

The community-wide BioLaySumm shared task initiative focuses on automatic biomedical lay summarisation and has attracted many researchers since 2023. The BioLaySumm shared task series encourages participants to develop and evaluate their systems within a common framework, and to compare their performance with other approaches \cite{shardlow2023biolaysumm,goldsack-etal-2024-biolaysumm,xiao-etal-2025-biolaysumm}. The shared task comprised two subtasks, standard lay summarisation and summarisation with external knowledge, and introduced the PLOS and eLife datasets, which pair biomedical articles with written lay summaries. Although both data sets were curated from full-text journal articles, they differ in authorship and length: PLOS uses short, author-written lay summaries, while eLife is based on longer summaries, approximately twice the length, written by expert editors.\cite{goldsack-etal-2022-making}.

The length of the text increases the computational cost of training and inference, as longer articles require more tokens to be processed \cite{hu2021lora,tay2022efficient}. In automatic lay summarisation, this aspect can become more critical, depending on the content provided as input to the model. For instance, an eLife abstract requires an average of 217 tokens, whereas the corresponding full-text article requires an average of 14,046 tokens, as shown in Table~\ref{tab:dataset-token-stats}.

The existing literature demonstrates considerable variation in how much source information is provided to biomedical lay summarization models. Some systems rely exclusively on abstracts \cite{turbitt-etal-2023-mdc, kim-etal-2024-saama-technologies, goldsack2025leveraging, lossio-ventura-etal-2025-5cnlp}, whereas others use abstracts supplemented with retrieved full-text passages \cite{lin-yu-2025-laysummx}, abstracts combined with conclusions \cite{bechler-etal-2025-tlpiq}, selected sections  \cite{wilson-etal-2025-raincitynlp}, salient sentences \cite{evgin2025metninozu, savasli2026graphxtract}, or the complete article \cite{liu-etal-2023-lhs712ee, colak-karadeniz-2023-isiksumm}. In addition, studies differ in their model architectures, prompting strategies, fine-tuning procedures, preprocessing methods, and evaluation settings. Consequently, direct conclusions about the effect of input scope remain difficult to draw.

Although several studies have compared or combined different input configurations, these comparisons are often embedded within different modeling frameworks or evaluated under different experimental conditions. For example, abstract-based methods may employ few-shot prompting \cite{turbitt-etal-2023-mdc, kim-etal-2024-saama-technologies, goldsack2025leveraging, lossio-ventura-etal-2025-5cnlp}, whereas full-text approaches may rely on fine-tuning or long-document architectures\cite{liu-etal-2023-lhs712ee, colak-karadeniz-2023-isiksumm}. Similarly, content-selection systems \cite{wilson-etal-2025-raincitynlp, evgin2025metninozu} introduce additional variables, which are the qualities of the selected sentences and the dependents of the extraction method. These differences make it difficult to determine whether improvements arise from the additional source information itself or from the particular model or pre-processing strategy used to exploit it. 

A particularly relevant question is therefore how much of a biomedical article is actually necessary for high-quality lay summarization. An abstract provides a concise representation of the article while it may omit methodological details, findings, or contextual information that could improve factuality and informativeness. Conversely, full-text input provides substantially more information, on the other hand it increases sequence length, computational cost, and the amount of potentially irrelevant content that the model must process. Although selected-content approaches attempt to address this trade-off, they introduce an additional content-selection component. 

The present work investigates the input-scope question explained above by systematically examining different amounts of source information under controlled experimental conditions. Rather than treating abstract-based and full-text summarization as fundamentally different tasks, we investigate them as alternative input configurations for the same general summarization objective. This enables a more direct analysis of the extent to which additional article-level information contributes to the quality of biomedical lay summarization, while also considering the associated information. In this study, we reported the experimental results on two different datasets with two different models and tested the statistical significance of the experimental results. Interestingly, our results show the opposite trend, with abstract-based approaches performing better than full-text approaches.

\noindent Based on these issues, this study addresses two research questions:
\begin{itemize}
    \item \textbf{RQ1:} Is an abstract alone sufficient to generate lay summaries that are comparable to those generated from the full article in terms of relevance, readability, and factual consistency?
    \item \textbf{RQ2:} Can few-shot prompting achieve performance comparable to task-specific LoRA fine-tuning while requiring less computational time?
\end{itemize}

\noindent The hypotheses of this study are as follows:
\begin{enumerate}
    \item \textbf{H1:} Abstract-based LORA fine-tuning achieves performance comparable to full-text LORA fine-tuning while requiring less computational time.

    \item \textbf{H2:} Three-shot prompting using abstract–summary example pairs achieves better overall performance compared with full-text LoRA fine-tuning while requiring less computational time.

    \item \textbf{H3:} Without fine-tuning, few-shot prompting with abstracts achieves better overall performance compared with few-shot prompting with full-text inputs while requiring less computational time.
\end{enumerate}

The main contribution of this study is a systematic, controlled evaluation of the sufficiency of abstracts for biomedical lay summarization, examining the trade-off between summary quality and computational efficiency. The results demonstrate that abstract-based LoRA fine-tuning achieves superior overall performance, as measured by a composite score integrating readability, relevance, and factual consistency, while reducing fine-tuning runtime by a factor of 4.70 compared with full-text LoRA fine-tuning. These findings challenge the assumption that access to full-text articles is necessary to produce high-quality biomedical lay summaries and suggest that abstracts may provide a more computationally efficient alternative without compromising, and potentially improving, overall summary quality.

Beyond fine-tuning, this study systematically investigates few-shot prompting with abstracts, broadening the comparison to include a parameter-efficient training approach and a training-free alternative. Furthermore, it expands the evaluation framework used in previous BioLaySumm studies by incorporating FENICE \cite{fenice2024} and LongDocFACTScore \cite{longdocfact2024} alongside established evaluation metrics. This multidimensional assessment provides a more comprehensive basis for comparing summarization strategies in terms of linguistic quality, relevance, factual consistency, and computational cost.

The remainder of this paper is organized as follows: Related Work, Methods, Results, Discussion, and Conclusion.

\section*{Related work}

Automatic summarization has been extensively studied across news, scientific, and biomedical domains. Scientific summarization presents particular challenges due to the length of research articles, the complexity of their content, and the need to preserve essential information \cite{cohan2018discourse}. To address the computational demands of long-document processing, models such as Longformer and Longformer Encoder-Decoder (LED) were developed to handle longer input sequences \cite{beltagy2020longformer}, while PEGASUS was designed specifically for abstractive summarization \cite{zhang2020pegasus}. Biomedical lay summarization introduces an additional challenge: generated summaries must convey scientifically accurate information in language accessible to non-expert readers. An important methodological consideration is therefore how much of the source article is needed to generate summaries that balance relevance, readability, and factual consistency. Accordingly, this section reviews previous work from three perspectives based on input strategy: abstract-based methods, content-selection methods, and full-text methods.

\subsection*{Abstract-based strategies}

Several studies have compared abstract input with longer article input for biomedical lay summarization. Basu et al. compared abstract-only and full-text fine-tuning on the eLife validation set and found trade-offs between relevance, readability, factuality, and inference time \cite{basu2025suwmit}. However, the comparison was limited to a single eLife split, so it remains unclear whether the same pattern holds on PLOS. Goldsack et al. reported a similar comparison for zero-shot prompting, again on eLife \cite{goldsack2025leveraging}. Together, these studies suggest that abstracts can be competitive with full text, but the evidence remains limited across datasets and adaptation methods.

Other studies extended abstract-based prompting to both PLOS and eLife. LaySummX compared abstract-only and full-text zero-shot prompting and found that abstract input gave better factuality scores on both datasets, while relevance and readability varied between them \cite{lin-yu-2025-laysummx}. The 5cNLP team also tested zero- and few-shot prompting with abstracts and different article sections \cite{lossio-ventura-etal-2025-5cnlp}. These studies covered a wider range of prompting settings, but did not compare the same abstract and full-text conditions across both prompting and fine-tuning.

Several studies have combined abstract-based fine-tuning with prompting. Saama Technologies used LoRA fine-tuning on abstracts together with zero- and six-shot generation \cite{kim-etal-2024-saama-technologies}, while SINAI compared two-shot prompting with abstract-based fine-tuning \cite{chizhikova-etal-2024-sinai}. DeakinNLP also compared fine-tuning with zero- and few-shot prompting, although its fine-tuning input included both abstracts and conclusions \cite{to-etal-2024-deakinnlp}. Earlier work by MDC similarly compared abstract-based fine-tuning with three-shot prompting \cite{turbitt-etal-2023-mdc}. These studies compared several adaptation methods, but did not apply the same methods to both abstract and full-text input.

Other work reduced the amount of article text used during fine-tuning. TLPIQ trained on the abstract and conclusion sections of PLOS and eLife \cite{bechler-etal-2025-tlpiq}, while MetninOzU limited each article to its first ten sentences \cite{evgin2025metninozu}. Both approaches showed that shorter inputs can support competitive lay summarization, but neither directly compared abstract-only and full-text fine-tuning under the same conditions.

\subsection*{Content selection strategies}

Several studies have explored strategies for reducing the amount of source text processed during lay summary generation. RainCityNLP employed TF–IDF to select a subset of sentences from full-text articles prior to training summarization models \cite{sparck1972tfidf,wilson-etal-2025-raincitynlp}. Similarly, MIRAGES adopted an extract-then-summarize approach, selecting salient sentences from full-text articles and fine-tuning Llama-3-8B-Instruct using low-rank adaptation (LoRA) \cite{pong-etal-2025-mirages}. These approaches aim to reduce the volume of input text processed by the summarization model while retaining relevant information from the full article.

Other systems have adopted alternative strategies to reduce the input length before summary generation. UIUC BioNLP extracted salient content from articles before fine-tuning GPT-3.5 \cite{you-etal-2024-uiuc}, whereas BDA-UC3M used extracted content in conjunction with compact models on the PLOS and eLife datasets \cite{ramzi-bedmar-2025-bda}. In contrast, Aard divided full-text articles into smaller chunks before performing zero-shot generation \cite{gupta-krishnamurthy-2025-shared}. Although these methods differ in their implementation, they share the objective of avoiding the processing of an entire article as a single input.

More recent work has investigated content selection explicitly as a means of improving summarization efficiency. Savaşlı et al. proposed GraphXtract, a graph-based approach that represents biomedical articles as graphs and selects salient sentences before LLM-based summary generation \cite{savasli2026graphxtract}. Evaluated on the PLOS dataset, the method was designed to reduce the volume of source text processed during summarization. Collectively, these approaches offer an alternative to strategies that rely exclusively on abstracts or process entire articles, enabling summarization models to operate on selected portions of the source text that are considered most informative.

\subsection*{Full-text-based strategies}

Full-text input has also been explored in biomedical lay summarization. Early BioLaySumm systems investigated long-document models, including BART and LED, to process full-text articles \cite{liu-etal-2023-lhs712ee}. Similarly, ISIKSumm employed text chunking to accommodate article length and generated separate summaries using abstract and full-text inputs \cite{colak-karadeniz-2023-isiksumm}. These studies explored strategies for incorporating information from longer biomedical documents, although the extent to which input scope affects summarization performance was not the primary focus.

CTyun AI adopted a related approach, using truncation and chunking to prepare full-text articles for fine-tuning \cite{bao-etal-2024-ctyun}. Rather than processing each article as a single sequence, the system shortened or divided the input to accommodate the constraints of long-document processing. Although these strategies offer practical solutions for handling lengthy biomedical texts, the study did not include a matched abstract-only condition to directly assess the effect of input scope.

More recent research has investigated the potential of smaller language models for biomedical lay summarization. Nursal and Mitchell evaluated models with fewer than 8 billion parameters in a zero-shot setting using full-text articles \cite{nursal-mitchell-2026-understanding}. Their findings indicated that smaller models could generate informative lay summaries, with some outputs also receiving favorable human evaluations. However, their evaluation did not extend to abstract-only input, few-shot prompting, or fine-tuning. Consequently, the relative effectiveness of different input scopes and adaptation strategies remains insufficiently characterized across these approaches.

\section*{Methods}

\begin{figure*}[!htp]
    \centering
    \includegraphics[scale=0.2]{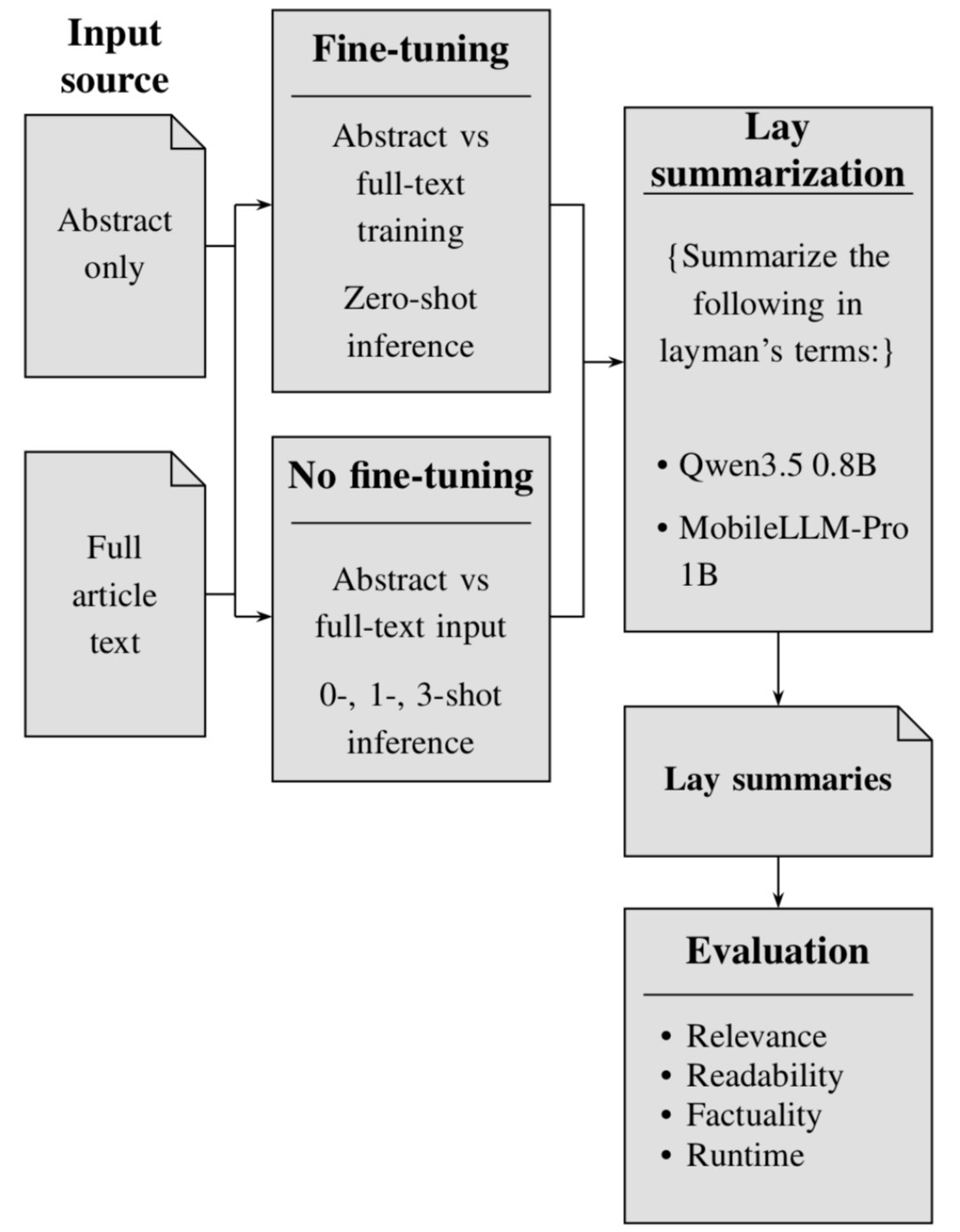}
    \caption{\textbf{Overall workflow of the study.} Biomedical articles from the PLOS and eLife datasets were used in abstract-based and full-text settings. The study compared LoRA fine-tuning and zero-/few-shot prompting, followed by evaluation using relevance, readability, factuality, and computational efficiency metrics.}
    \label{fig:workflow}
\end{figure*}

The methods are organized into three parts: dataset, fine-tuning and inference, and evaluation. Fig~\ref{fig:workflow} shows the overall workflow of the study.

\subsection*{Dataset}

The PLOS and eLife datasets consist of biomedical research articles paired with expert-written lay summaries \cite{goldsack-etal-2022-making}. During preprocessing, abstracts were extracted from the article text and used as input for the abstract-only condition. The datasets were obtained from BioLaySumm 2025 \cite{xiao-etal-2025-biolaysumm}. Since gold-standard lay summaries were unavailable for the original test sets, new test sets were sampled from the training data. Each new test set comprised approximately 5\% of the corresponding original training set, yielding a split close to 90/5/5 for training, validation, and testing, respectively (Table~\ref{tab:dataset-splits}).

\begin{table}[!ht]
\centering
\caption{\textbf{Dataset split sizes for eLife and PLOS.}}
\label{tab:dataset-splits}
\resizebox{\textwidth}{!}{%
\begin{tabular}{lrrrrr}
\hline
Dataset & Train & Validation & Test & Prompt set & Few-shot samples \\
\hline
eLife & 4,066 (89.68\%) & 241 (5.32\%) & 227 (5.01\%) & 25 & 3 \\
PLOS & 23,413 (89.73\%) & 1,376 (5.27\%) & 1,305 (5.00\%) & 25 & 3 \\
\hline
\end{tabular}%
}
\end{table}

All sampling was performed before the experiments using a fixed random seed of 42, without stratification. The newly constructed test sets, the prompt-selection set, and the few-shot samples were held out from model training and fine-tuning to prevent data leakage. The prompt-selection set comprised 25 samples from each dataset, while three samples from each dataset were used for few-shot prompting. The number of few-shot samples was limited to three because some eLife articles exceed 35,000 tokens, potentially exceeding the model's context-length limit.

\begin{table}[!ht]
\centering
\caption{\textbf{Token statistics for eLife and PLOS.}}
\label{tab:dataset-token-stats}
\resizebox{\textwidth}{!}{%
\begin{tabular}{lrrrrr}
\hline
Dataset & Avg. article tokens & Max. article tokens &
Min. article tokens & Avg. abstract tokens &
Abstract--article token overlap \\
\hline
eLife & 14{,}046.1 & 38{,}940 & 3{,}020 & 217.39 & 21\% \\
PLOS & 8{,}902.0 & 24{,}522 & 1{,}195 & 348.74 & 49\% \\
\hline
\end{tabular}%
}

\begin{flushleft}
Abstract--article token overlap was calculated for each article from the tokens shared by the abstract and the full article, and then averaged for each dataset. Token counts were obtained using the tokenizer provided with the \texttt{Qwen/Qwen3.5-0.8B} checkpoint.
\end{flushleft}
\end{table}

As reported in previous work \cite{goldsack-etal-2022-making}, PLOS articles are shorter on average than eLife articles, a difference also reflected in the token statistics reported in Table~\ref{tab:dataset-token-stats}.

\subsection*{Fine-tuning and inference}

The experiments were conducted using Qwen3.5-0.8B \cite{qwen3.5} and MobileLLM-Pro-1B \cite{mobilellm}. These models were selected because their context lengths were sufficient for the articles in the datasets and their computational requirements were compatible with the available hardware. All experiments were conducted in the same Google Colab environment using an NVIDIA L4 GPU.

\textbf{Prompt selection} Ten prompt formulations were evaluated on the prompt-selection set using both models with greedy decoding. The prompt with the highest mean score across the same 11 evaluation metrics used in the main experiments was selected for subsequent experiments. The selected prompt was ``Summarize the following in layman's terms:'' and was used throughout the remaining experiments. The complete set of prompt formulations and the corresponding prompt-selection results are provided in the Supplementary Material.

\textbf{Fine-tuning configuration} All fine-tuning experiments used identical Low-Rank Adaptation (LoRA) settings. LoRA was applied to the \texttt{q\_proj} and \texttt{v\_proj} layers, with a rank ((r)) of 4, an alpha value of 8, and a dropout rate of 0.05. The rank was set to 4 in accordance with the original LoRA study, which demonstrated competitive performance using low-rank adaptation of query and value projections \cite{hu2021lora}. Models were trained for two epochs using a learning rate of ($2 \times 10^{-4}$), a batch size of 2, and eight gradient accumulation steps. Weight decay was set to 0.01, with 50 warm-up steps. Training was performed using \texttt{bfloat16} precision, gradient checkpointing, and scaled dot-product attention (SDPA).

Fine-tuning was conducted under two input conditions: full-article input and abstract-only input. As reported in Table~\ref{tab:dataset-token-stats}, abstracts contain approximately 25.5 times fewer tokens than full articles in PLOS and 64.6 times fewer in eLife. Each fine-tuning condition was trained once, starting from the same base model, without intermediate checkpoint selection.

\textbf{Inference configuration} The maximum number of newly generated tokens was set to 800 for all summaries. This limit was chosen to exceed the average lay-summary lengths reported for both PLOS and eLife \cite{goldsack-etal-2022-making}. Sampling was disabled using \texttt{do\_sample=False}, and generation employed greedy decoding. 

Consequently, the temperature, \texttt{top\_k}, and \texttt{top\_p} parameters did not affect generation. The end-of-sequence token was also used as the padding token. Identical inference settings were applied to zero-shot, one-shot, three-shot, and fine-tuned generation.

\subsection*{Evaluation}

The generated summaries were evaluated along three dimensions: relevance, readability, and factuality. Relevance was assessed using ROUGE \cite{lin2004rouge}, BLEU \cite{papineni2002bleu}, METEOR \cite{banerjee2005meteor}, and BERTScore \cite{zhang2019bertscore}. ROUGE, BLEU, and METEOR quantify lexical similarity between generated summaries and their reference summaries, whereas BERTScore assesses semantic similarity using contextual embeddings. Readability was evaluated using the Flesch–Kincaid Grade Level (FKGL) \cite{kincaid1975derivation}, the Dale–Chall Readability Score (DCRS) \cite{chall1995readability}, the SMOG index \cite{mc1969smog}, and the Coleman–Liau Index (CLI) \cite{coleman1975computer}. Lower values indicate greater readability, or easier-to-read text, for all four metrics. Factual consistency was assessed using SummaC \cite{laban2021summac}, FENICE \cite{fenice2024}, and LongDocFactScore \cite{longdocfact2024}. These metrics assess the consistency of generated content with the source text, providing complementary signals of potential factual errors.

For the normalization and aggregation of these three metrics,  no explicit weighting was assigned to the three evaluation dimensions. Instead, all 11 metrics contributed equally to the overall score. Consequently, relevance and readability each accounted for (4/11) (36.36\%) of the aggregate, while factuality accounted for (3/11) (27.27\%).
Before aggregation, each metric was normalized using min–max scaling. For metrics in which higher values indicate better performance, normalization was performed as follows:

\begin{equation}
\label{eq:normalization-higher}
x' = \frac{x - \min(x)}{\max(x) - \min(x)}.
\end{equation}
For metrics in which lower values indicate better performance, the normalized score was calculated as:
\begin{equation}
\label{eq:normalization-lower}
x' = 1 - \frac{x - \min(x)}{\max(x) - \min(x)}.
\end{equation}

These transformations map the metric values to a common scale on which higher normalized scores consistently indicate better performance. The overall score was calculated as the arithmetic mean of the 11 normalized metric scores.

A limited manual assessment was conducted to examine potential hallucinations in outputs with the lowest factuality scores. These outputs were inspected for unsupported or contradictory claims. The observations were consistent with the low factuality scores assigned to these outputs.

For statistical analysis, each evaluation metric was calculated for every prediction in the test set. Statistical significance between paired experimental conditions was assessed using a two-sided Wilcoxon signed-rank test \cite{wilcoxon1945}, with scores paired by the corresponding test articles. The significance threshold was set to ($\alpha=0.05$). The rank-biserial correlation was reported as the effect size.

\section*{Results}

The results are presented according to the run time and the three hypotheses.

\subsection*{Runtime}

Table~\ref{tab:runtimes-hms} shows the total runtimes of each setting across the two models and two datasets. All experiments were run in the same Google Colab environment with an NVIDIA L4 GPU.

\begin{table}[!ht]
\centering
\caption{\textbf{Total runtimes of each setting across two models and two datasets.}}
\label{tab:runtimes-hms}
\resizebox{\textwidth}{!}{%
\begin{tabular}{lllrrr}
\toprule
\textbf{Dataset} &
\textbf{Setting} &
\textbf{Model} &
\textbf{Fine-tuning duration} &
\textbf{Inference duration} &
\textbf{Total duration} \\
\midrule

eLife & Abstract Zero-Shot
& Qwen3.5-0.8B
& -- & 28 min 41 s & 28 min 41 s \\

eLife & Abstract One-Shot
& Qwen3.5-0.8B
& -- & 35 min 12 s & 35 min 12 s \\

eLife & Abstract Three-Shot
& Qwen3.5-0.8B
& -- & 48 min 58 s & 48 min 58 s \\

eLife & Full-Text Zero-Shot
& Qwen3.5-0.8B
& -- & 1 h 19 min 11 s & 1 h 19 min 11 s \\

eLife & Full-Text One-Shot
& Qwen3.5-0.8B
& -- & 1 h 31 min 56 s & 1 h 31 min 56 s \\

eLife & Full-Text Three-Shot
& Qwen3.5-0.8B
& -- & 1 h 50 min 04 s & 1 h 50 min 04 s \\

\addlinespace

eLife & Abstract Zero-Shot (Fine-Tuned)
& Qwen3.5-0.8B
& 24 min 15 s & 1 h 19 min 00 s & 1 h 43 min 15 s \\

eLife & Full-Text Zero-Shot (Fine-Tuned)
& Qwen3.5-0.8B
& 2 h 48 min 04 s & 5 h 29 min 00 s & 8 h 17 min 04 s \\

\addlinespace

eLife & Abstract Zero-Shot
& MobileLLM-Pro (1B)
& -- & 30 min 21 s & 30 min 21 s \\

eLife & Abstract One-Shot
& MobileLLM-Pro (1B)
& -- & 44 min 34 s & 44 min 34 s \\

eLife & Abstract Three-Shot
& MobileLLM-Pro (1B)
& -- & 56 min 04 s & 56 min 04 s \\

eLife & Full-Text Zero-Shot
& MobileLLM-Pro (1B)
& -- & 1 h 28 min 54 s & 1 h 28 min 54 s \\

eLife & Full-Text One-Shot
& MobileLLM-Pro (1B)
& -- & 1 h 46 min 42 s & 1 h 46 min 42 s \\

eLife & Full-Text Three-Shot
& MobileLLM-Pro (1B)
& -- & 2 h 08 min 51 s & 2 h 08 min 51 s \\

\addlinespace

eLife & Abstract Zero-Shot (Fine-Tuned)
& MobileLLM-Pro (1B)
& 28 min 48 s & 1 h 24 min 57 s & 1 h 53 min 45 s \\

eLife & Full-Text Zero-Shot (Fine-Tuned)
& MobileLLM-Pro (1B)
& 2 h 55 min 10 s & 5 h 36 min 46 s & 8 h 31 min 56 s \\

\midrule

PLOS & Abstract Zero-Shot
& Qwen3.5-0.8B
& -- & 3 h 03 min 41 s & 3 h 03 min 41 s \\

PLOS & Abstract One-Shot
& Qwen3.5-0.8B
& -- & 4 h 43 min 35 s & 4 h 43 min 35 s \\

PLOS & Abstract Three-Shot
& Qwen3.5-0.8B
& -- & 5 h 51 min 01 s & 5 h 51 min 01 s \\

PLOS & Full-Text Zero-Shot
& Qwen3.5-0.8B
& -- & 34 h 41 min 09 s & 34 h 41 min 09 s \\

PLOS & Full-Text One-Shot
& Qwen3.5-0.8B
& -- & 37 h 42 min 30 s & 37 h 42 min 30 s \\

PLOS & Full-Text Three-Shot
& Qwen3.5-0.8B
& -- & 40 h 35 min 23 s & 40 h 35 min 23 s \\

\addlinespace

PLOS & Abstract Zero-Shot (Fine-Tuned)
& Qwen3.5-0.8B
& 2 h 25 min 51 s & 6 h 41 min 00 s & 9 h 06 min 51 s \\

PLOS & Full-Text Zero-Shot (Fine-Tuned)
& Qwen3.5-0.8B
& 16 h 32 min 43 s & 28 h 05 min 00 s & 44 h 37 min 43 s \\

\addlinespace

PLOS & Abstract Zero-Shot
& MobileLLM-Pro (1B)
& -- & 3 h 39 min 18 s & 3 h 39 min 18 s \\

PLOS & Abstract One-Shot
& MobileLLM-Pro (1B)
& -- & 5 h 06 min 04 s & 5 h 06 min 04 s \\

PLOS & Abstract Three-Shot
& MobileLLM-Pro (1B)
& -- & 6 h 17 min 07 s & 6 h 17 min 07 s \\

PLOS & Full-Text Zero-Shot
& MobileLLM-Pro (1B)
& -- & 36 h 01 min 52 s & 36 h 01 min 52 s \\

PLOS & Full-Text One-Shot
& MobileLLM-Pro (1B)
& -- & 39 h 23 min 51 s & 39 h 23 min 51 s \\

PLOS & Full-Text Three-Shot
& MobileLLM-Pro (1B)
& -- & 44 h 53 min 21 s & 44 h 53 min 21 s \\

\addlinespace

PLOS & Abstract Zero-Shot (Fine-Tuned)
& MobileLLM-Pro (1B)
& 2 h 37 min 29 s & 7 h 34 min 50 s & 10 h 12 min 19 s \\

PLOS & Full-Text Zero-Shot (Fine-Tuned)
& MobileLLM-Pro (1B)
& 17 h 14 min 31 s & 29 h 17 min 58 s & 46 h 32 min 29 s \\

\bottomrule
\end{tabular}%
}

\begin{flushleft}
\footnotesize
All experiments used a Google Colab L4 GPU with 24 GB of VRAM and the same parameter settings.
\end{flushleft}
\end{table}

\subsection*{H1: Abstract-based vs. full-text LoRA fine-tuning}

H1 examines the effects of input scope on summary quality and fine-tuning runtime by comparing abstract-based and full-text LoRA fine-tuning. The comparison was conducted separately on the eLife and PLOS datasets using both models.

\subsubsection*{H1: PLOS}

Table~\ref{tab:plos-fine-tuning-results} presents the results of abstract-based and full-text LoRA fine-tuning on PLOS. Both models achieved higher overall scores with abstract-based fine-tuning than with full-text fine-tuning (Qwen3.5-0.8B: Holm-adjusted $p<.001$, $r_{rb}=.68$; MobileLLM-Pro-1B: Holm-adjusted $p<.001$, $r_{rb}=.76$).

\begin{table}[!ht]
\centering
\caption{\textbf{PLOS fine-tuning results on the test set.}}
\label{tab:plos-fine-tuning-results}
\resizebox{\textwidth}{!}{%
\begin{tabular}{@{}l*{12}{c}@{}}
\hline
& \multicolumn{4}{c}{\textbf{Relevance}}
& \multicolumn{4}{c}{\textbf{Readability}}
& \multicolumn{3}{c}{\textbf{Factuality}}
& \\
\cline{2-5}
\cline{6-9}
\cline{10-12}
\rotatebox{90}{\textbf{Setting}}
& \rotatebox{90}{\textbf{ROUGE\rotatebox{-90}{$\uparrow$}}}
& \rotatebox{90}{\textbf{BLEU\rotatebox{-90}{$\uparrow$}}}
& \rotatebox{90}{\textbf{METEOR\rotatebox{-90}{$\uparrow$}}}
& \rotatebox{90}{\textbf{BERTScore\rotatebox{-90}{$\uparrow$}}}
& \rotatebox{90}{\textbf{FKGL\rotatebox{-90}{$\downarrow$}}}
& \rotatebox{90}{\textbf{DCRS\rotatebox{-90}{$\downarrow$}}}
& \rotatebox{90}{\textbf{SMOG\rotatebox{-90}{$\downarrow$}}}
& \rotatebox{90}{\textbf{CLI\rotatebox{-90}{$\downarrow$}}}
& \rotatebox{90}{\textbf{SummaC\rotatebox{-90}{$\uparrow$}}}
& \rotatebox{90}{\textbf{FENICE\rotatebox{-90}{$\uparrow$}}}
& \rotatebox{90}{\textbf{LongDocFact\rotatebox{-90}{$\uparrow$}}}
& \rotatebox{90}{\textbf{Mean\rotatebox{-90}{$\uparrow$}}} \\
\hline
MobileLLM-Pro Abstract Fine-Tuning
& 0.4052 & \textbf{0.1238} & 0.2257 & 0.8583
& 14.71 & 12.37 & \textbf{14.49} & \textbf{14.03}
& 0.7654 & 0.9256 & 0.6234 & \textbf{0.7840} \\

Qwen Abstract Fine-Tuning
& 0.2699 & 0.1119 & \textbf{0.2999} & \textbf{0.8633}
& \textbf{13.86} & \textbf{11.77} & 14.89 & 17.11
& 0.7236 & 0.9249 & \textbf{0.7262} & 0.7262 \\

Qwen Full-Text Fine-Tuning
& \textbf{0.4077} & 0.1197 & 0.1921 & 0.8448
& 14.66 & 13.39 & 15.34 & 15.95
& \textbf{0.7863} & \textbf{0.9482} & 0.5743 & 0.5526 \\

MobileLLM-Pro Full-Text Fine-Tuning
& 0.0854 & 0.0090 & 0.1410 & 0.7919
& 16.31 & 11.79 & 15.19 & 15.35
& 0.7125 & 0.8067 & 0.5897 & 0.1671 \\
\hline
\end{tabular}%
}
\end{table}

Abstract-based fine-tuning also yielded lower scores on three of the four readability metrics for both models, indicating improvements in readability on those measures. As reported in Table~\ref{tab:runtimes-hms}, abstract-based fine-tuning was 4.90 times faster for Qwen3.5-0.8B and 4.56 times faster for MobileLLM-Pro-1B than full-text fine-tuning. These findings support H1 on PLOS in terms of both overall summary quality and runtime.

\subsubsection*{H1: eLife}

Table~\ref{tab:elife-fine-tuning-results} presents the corresponding results for eLife. As on PLOS, abstract-based fine-tuning achieved higher overall scores than full-text fine-tuning for both models (Qwen3.5-0.8B: Holm-adjusted $p=.004$, $r_{rb}=.62$; MobileLLM-Pro-1B: Holm-adjusted $p<.001$, $r_{rb}=.81$). Among the reported eLife fine-tuning conditions, abstract-based fine-tuning with MobileLLM-Pro-1B achieved the highest overall score. Under full-text fine-tuning, Qwen3.5-0.8B achieved a higher overall score than MobileLLM-Pro-1B.

\begin{table}[!ht]
\centering
\caption{\textbf{eLife fine-tuning results on the test set.}}
\label{tab:elife-fine-tuning-results}
\resizebox{\textwidth}{!}{%
\begin{tabular}{@{}l*{12}{c}@{}}
\hline
& \multicolumn{4}{c}{\textbf{Relevance}}
& \multicolumn{4}{c}{\textbf{Readability}}
& \multicolumn{3}{c}{\textbf{Factuality}}
& \\
\cline{2-5}
\cline{6-9}
\cline{10-12}
\rotatebox{90}{\textbf{Setting}}
& \rotatebox{90}{\textbf{ROUGE\rotatebox{-90}{$\uparrow$}}}
& \rotatebox{90}{\textbf{BLEU\rotatebox{-90}{$\uparrow$}}}
& \rotatebox{90}{\textbf{METEOR\rotatebox{-90}{$\uparrow$}}}
& \rotatebox{90}{\textbf{BERTScore\rotatebox{-90}{$\uparrow$}}}
& \rotatebox{90}{\textbf{FKGL\rotatebox{-90}{$\downarrow$}}}
& \rotatebox{90}{\textbf{DCRS\rotatebox{-90}{$\downarrow$}}}
& \rotatebox{90}{\textbf{SMOG\rotatebox{-90}{$\downarrow$}}}
& \rotatebox{90}{\textbf{CLI\rotatebox{-90}{$\downarrow$}}}
& \rotatebox{90}{\textbf{SummaC\rotatebox{-90}{$\uparrow$}}}
& \rotatebox{90}{\textbf{FENICE\rotatebox{-90}{$\uparrow$}}}
& \rotatebox{90}{\textbf{LongDocFact\rotatebox{-90}{$\uparrow$}}}
& \rotatebox{90}{\textbf{Mean\rotatebox{-90}{$\uparrow$}}} \\
\hline
MobileLLM-Pro Abstract Fine-Tuning
& 0.3360 & \textbf{0.0797} & 0.1746 & 0.8378
& \textbf{12.21} & \textbf{10.99} & 12.20 & 14.38
& 0.8745 & 0.8634 & \textbf{0.6457} & \textbf{0.8149} \\

Qwen Abstract Fine-Tuning
& 0.2357 & 0.0505 & \textbf{0.2482} & \textbf{0.8456}
& 13.27 & 11.99 & 15.00 & \textbf{13.70}
& 0.6669 & 0.7142 & 0.6295 & 0.5780 \\

Qwen Full-Text Fine-Tuning
& \textbf{0.3483} & 0.0705 & 0.1638 & 0.8269
& 15.69 & 13.62 & 16.66 & 16.66
& \textbf{0.9911} & \textbf{0.9594} & 0.6001 & 0.4377 \\

MobileLLM-Pro Full-Text Fine-Tuning
& 0.1098 & 0.0095 & 0.1331 & 0.7877
& 14.89 & 12.01 & \textbf{10.85} & 14.01
& 0.9032 & 0.8423 & 0.6055 & 0.3732 \\
\hline
\end{tabular}%
}

\begin{flushleft}
\footnotesize
Qwen refers to Qwen3.5-0.8B, and MobileLLM-Pro refers to the MobileLLM-Pro-1B model.
\end{flushleft}
\end{table}

The runtime results in Table~\ref{tab:runtimes-hms} show that abstract-based fine-tuning was 4.83 times faster for Qwen3.5-0.8B and 4.49 times faster for MobileLLM-Pro-1B than full-text fine-tuning. Thus, the eLife results also support H1 in terms of overall summary quality and runtime.

\subsection*{H2: Three-shot abstract prompting vs. full-text LoRA fine-tuning}

H2 compares three-shot abstract prompting with full-text LoRA fine-tuning in terms of summary quality and runtime. The comparison was conducted separately on the eLife and PLOS datasets using both models.

\subsubsection*{H2: PLOS}

Table~\ref{tab:plos-best-settings} presents the results for the best-performing prompting and fine-tuning strategies on PLOS.

\begin{table}[!ht]
\centering
\caption{\textbf{Best settings for each model on the PLOS test set.}}
\label{tab:plos-best-settings}
\resizebox{\textwidth}{!}{%
\begin{tabular}{@{}l*{12}{c}@{}}
\hline
& \multicolumn{4}{c}{\textbf{Relevance}}
& \multicolumn{4}{c}{\textbf{Readability}}
& \multicolumn{3}{c}{\textbf{Factuality}}
& \\
\cline{2-5}
\cline{6-9}
\cline{10-12}
\rotatebox{90}{\textbf{Setting}}
& \rotatebox{90}{\textbf{ROUGE\rotatebox{-90}{$\uparrow$}}}
& \rotatebox{90}{\textbf{BLEU\rotatebox{-90}{$\uparrow$}}}
& \rotatebox{90}{\textbf{METEOR\rotatebox{-90}{$\uparrow$}}}
& \rotatebox{90}{\textbf{BERTScore\rotatebox{-90}{$\uparrow$}}}
& \rotatebox{90}{\textbf{FKGL\rotatebox{-90}{$\downarrow$}}}
& \rotatebox{90}{\textbf{DCRS\rotatebox{-90}{$\downarrow$}}}
& \rotatebox{90}{\textbf{SMOG\rotatebox{-90}{$\downarrow$}}}
& \rotatebox{90}{\textbf{CLI\rotatebox{-90}{$\downarrow$}}}
& \rotatebox{90}{\textbf{SummaC\rotatebox{-90}{$\uparrow$}}}
& \rotatebox{90}{\textbf{FENICE\rotatebox{-90}{$\uparrow$}}}
& \rotatebox{90}{\textbf{LongDocFact\rotatebox{-90}{$\uparrow$}}}
& \rotatebox{90}{\textbf{Mean\rotatebox{-90}{$\uparrow$}}} \\
\hline
Qwen 3-Shot Abstract
& 0.3471 & 0.0649 & 0.1678 & 0.8413
& \textbf{10.49} & \textbf{10.63} & \textbf{10.67} & \textbf{12.79}
& 0.7712 & 0.8808 & 0.5930 & \textbf{0.6588} \\

MobileLLM-Pro 3-Shot Abstract
& 0.3537 & 0.0812 & 0.1881 & 0.8417
& 13.14 & 11.31 & 12.37 & 14.98
& \textbf{0.8743} & 0.9246 & 0.5594 & 0.6108 \\

MobileLLM-Pro Abstract Fine-Tuning
& 0.4052 & \textbf{0.1238} & 0.2257 & 0.8583
& 14.71 & 12.37 & 14.49 & 14.03
& 0.7654 & 0.9256 & 0.6234 & 0.5954 \\

Qwen Abstract Fine-Tuning
& 0.2699 & 0.1119 & \textbf{0.2999} & \textbf{0.8633}
& 13.86 & 11.77 & 14.89 & 17.11
& 0.7236 & 0.9249 & \textbf{0.7262} & 0.5890 \\

Qwen Full-Text Fine-Tuning
& \textbf{0.4077} & 0.1197 & 0.1921 & 0.8448
& 14.66 & 13.39 & 15.34 & 15.95
& 0.7863 & \textbf{0.9482} & 0.5743 & 0.4659 \\

MobileLLM-Pro Full-Text Fine-Tuning
& 0.0854 & 0.0090 & 0.1410 & 0.7919
& 16.31 & 11.79 & 15.19 & 15.35
& 0.7125 & 0.8067 & 0.5897 & 0.1093 \\
\hline
\end{tabular}%
}

\begin{flushleft}
\footnotesize
Qwen refers to Qwen3.5-0.8B, and MobileLLM-Pro refers to the MobileLLM-Pro-1B model.
\end{flushleft}
\end{table}

Three-shot abstract prompting achieved higher overall scores than full-text fine-tuning for both models (Qwen3.5-0.8B: $p=.042$, $r_{rb}=.517$; MobileLLM-Pro-1B: $p<.001$, $r_{rb}=.69$). It also yielded lower scores on all four readability metrics for both models, indicating easier-to-read summaries across these measures.

In terms of runtime, three-shot abstract prompting was 7.63 times faster for Qwen3.5-0.8B and 7.41 times faster for MobileLLM-Pro-1B than full-text fine-tuning (Table~\ref{tab:runtimes-hms}). Moreover, three-shot abstract prompting achieved a higher overall mean score than abstract-based fine-tuning for both models. These results support H2 on PLOS in terms of overall summary quality and runtime.

\subsubsection*{H2: eLife}

Table~\ref{tab:elife-best-settings} presents the results for the best-performing prompting and fine-tuning strategies on eLife.

\begin{table}[!ht]
\centering
\caption{\textbf{Best settings for each model on the eLife test set.}}
\label{tab:elife-best-settings}
\resizebox{\textwidth}{!}{%
\begin{tabular}{@{}l*{12}{c}@{}}
\hline
& \multicolumn{4}{c}{\textbf{Relevance}}
& \multicolumn{4}{c}{\textbf{Readability}}
& \multicolumn{3}{c}{\textbf{Factuality}}
& \\
\cline{2-5}
\cline{6-9}
\cline{10-12}
\rotatebox{90}{\textbf{Setting}}
& \rotatebox{90}{\textbf{ROUGE\rotatebox{-90}{$\uparrow$}}}
& \rotatebox{90}{\textbf{BLEU\rotatebox{-90}{$\uparrow$}}}
& \rotatebox{90}{\textbf{METEOR\rotatebox{-90}{$\uparrow$}}}
& \rotatebox{90}{\textbf{BERTScore\rotatebox{-90}{$\uparrow$}}}
& \rotatebox{90}{\textbf{FKGL\rotatebox{-90}{$\downarrow$}}}
& \rotatebox{90}{\textbf{DCRS\rotatebox{-90}{$\downarrow$}}}
& \rotatebox{90}{\textbf{SMOG\rotatebox{-90}{$\downarrow$}}}
& \rotatebox{90}{\textbf{CLI\rotatebox{-90}{$\downarrow$}}}
& \rotatebox{90}{\textbf{SummaC\rotatebox{-90}{$\uparrow$}}}
& \rotatebox{90}{\textbf{FENICE\rotatebox{-90}{$\uparrow$}}}
& \rotatebox{90}{\textbf{LongDocFact\rotatebox{-90}{$\uparrow$}}}
& \rotatebox{90}{\textbf{Mean\rotatebox{-90}{$\uparrow$}}} \\
\hline
MobileLLM-Pro Abstract Fine-Tuning
& 0.3360 & \textbf{0.0797} & 0.1746 & 0.8378
& 12.21 & \textbf{10.99} & 12.20 & 14.38
& 0.8745 & 0.8634 & 0.6457 & \textbf{0.7472} \\

MobileLLM-Pro 3-Shot Abstract
& 0.2929 & 0.0632 & 0.1535 & 0.8297
& 11.90 & 11.83 & 13.04 & 15.65
& 0.9054 & 0.9145 & \textbf{0.6585} & 0.6585 \\

Qwen 3-Shot Abstract
& \textbf{0.3643} & 0.0605 & 0.1581 & 0.8373
& \textbf{10.56} & 11.06 & 11.75 & \textbf{12.69}
& 0.7938 & 0.6995 & 0.5901 & 0.6372 \\

Qwen Abstract Fine-Tuning
& 0.2357 & 0.0505 & \textbf{0.2482} & \textbf{0.8456}
& 13.27 & 11.99 & 15.00 & 13.70
& 0.6669 & 0.7142 & 0.6295 & 0.5303 \\

Qwen Full-Text Fine-Tuning
& 0.3483 & 0.0705 & 0.1638 & 0.8269
& 15.69 & 13.62 & 16.66 & 16.66
& \textbf{0.9911} & \textbf{0.9594} & 0.6001 & 0.4452 \\

MobileLLM-Pro Full-Text Fine-Tuning
& 0.1098 & 0.0095 & 0.1331 & 0.7877
& 14.89 & 12.01 & \textbf{10.85} & 14.01
& 0.9032 & 0.8423 & 0.6055 & 0.3579 \\
\hline
\end{tabular}%
}

\begin{flushleft}
\footnotesize
Qwen refers to Qwen3.5-0.8B, and MobileLLM-Pro refers to MobileLLM-Pro-1B.
\end{flushleft}
\end{table}

Three-shot abstract prompting achieved higher overall scores than full-text fine-tuning for both models (Qwen3.5-0.8B: $p<.001$, $r_{rb}=.95$; MobileLLM-Pro-1B: $p<.001$, $r_{rb}=.603$). However, the results for relevance, readability, and factuality varied across the two models, indicating that the overall advantage did not translate into a uniform improvement across all evaluation dimensions.

Three-shot abstract prompting was 10.14 times faster for Qwen3.5-0.8B and 9.14 times faster for MobileLLM-Pro-1B than full-text fine-tuning (Table~\ref{tab:runtimes-hms}). Compared with abstract-based fine-tuning, three-shot abstract prompting achieved a higher overall mean score for Qwen3.5-0.8B, whereas abstract-based fine-tuning achieved a higher overall mean score for MobileLLM-Pro-1B. Nevertheless, the comparison with full-text fine-tuning supports H2 on eLife in terms of overall summary quality and runtime.

\subsection*{H3: Abstract vs. full-text few-shot prompting}

H3 investigates the effect of input scope on few-shot summarization by comparing abstract-based and full-text prompting under zero-shot, one-shot, and three-shot settings. The comparisons were conducted separately on the eLife and PLOS datasets using both models, with summary quality and runtime as the primary outcomes.

\subsubsection*{H3: PLOS}

Table~\ref{tab:few-shot-plos} presents the few-shot inference results for both models on PLOS.

\begin{table}[!ht]
\centering
\caption{\textbf{Few-shot inference results on the PLOS test set.}}
\label{tab:few-shot-plos}
\resizebox{\textwidth}{!}{%
\begin{tabular}{@{}l*{12}{c}@{}}
\hline
& \multicolumn{4}{c}{\textbf{Relevance}}
& \multicolumn{4}{c}{\textbf{Readability}}
& \multicolumn{3}{c}{\textbf{Factuality}}
& \\
\cline{2-5}
\cline{6-9}
\cline{10-12}
\rotatebox{90}{\textbf{Setting}}
& \rotatebox{90}{\textbf{ROUGE\rotatebox{-90}{$\uparrow$}}}
& \rotatebox{90}{\textbf{BLEU\rotatebox{-90}{$\uparrow$}}}
& \rotatebox{90}{\textbf{METEOR\rotatebox{-90}{$\uparrow$}}}
& \rotatebox{90}{\textbf{BERTScore\rotatebox{-90}{$\uparrow$}}}
& \rotatebox{90}{\textbf{FKGL\rotatebox{-90}{$\downarrow$}}}
& \rotatebox{90}{\textbf{DCRS\rotatebox{-90}{$\downarrow$}}}
& \rotatebox{90}{\textbf{SMOG\rotatebox{-90}{$\downarrow$}}}
& \rotatebox{90}{\textbf{CLI\rotatebox{-90}{$\downarrow$}}}
& \rotatebox{90}{\textbf{SummaC\rotatebox{-90}{$\uparrow$}}}
& \rotatebox{90}{\textbf{FENICE\rotatebox{-90}{$\uparrow$}}}
& \rotatebox{90}{\textbf{LongDocFact\rotatebox{-90}{$\uparrow$}}}
& \rotatebox{90}{\textbf{Mean\rotatebox{-90}{$\uparrow$}}} \\
\hline
Qwen 3-Shot Abstract
& 0.3471 & 0.0649 & 0.1678 & 0.8413
& \textbf{10.49} & \textbf{10.63} & \textbf{10.67} & 12.79
& 0.7712 & 0.8808 & 0.5930 & \textbf{0.7366} \\

Qwen 1-Shot Abstract
& 0.3351 & 0.0700 & 0.1789 & 0.8426
& 10.73 & 10.73 & 10.74 & \textbf{12.65}
& 0.6628 & 0.8788 & \textbf{0.5965} & 0.7364 \\

Qwen 0-Shot Abstract
& 0.3508 & 0.0695 & 0.1828 & 0.8419
& 11.25 & 10.72 & 10.77 & 13.30
& 0.5204 & 0.8890 & 0.5953 & 0.7060 \\

MobileLLM-Pro 3-Shot Abstract
& 0.3537 & 0.0812 & 0.1881 & 0.8417
& 13.14 & 11.31 & 12.37 & 14.98
& 0.8743 & 0.9246 & 0.5594 & 0.6954 \\

MobileLLM-Pro 1-Shot Abstract
& 0.3497 & 0.0860 & 0.1909 & 0.8455
& 13.31 & 11.48 & 12.76 & 15.10
& 0.8545 & 0.9133 & 0.5657 & 0.6909 \\

Qwen 1-Shot
& 0.3583 & 0.0811 & 0.1707 & 0.8283
& 12.07 & 11.76 & 12.11 & 14.10
& \textbf{0.9724} & 0.9519 & 0.5133 & 0.6465 \\

Qwen 0-Shot
& 0.3199 & 0.0655 & 0.1558 & 0.8216
& 10.58 & 11.06 & 10.68 & 12.73
& 0.9257 & \textbf{0.9647} & 0.4832 & 0.6363 \\

Qwen 3-Shot
& \textbf{0.3696} & \textbf{0.1031} & 0.1777 & 0.8283
& 13.16 & 12.57 & 13.83 & 14.91
& 0.8036 & 0.9479 & 0.5197 & 0.5666 \\

MobileLLM-Pro 0-Shot Abstract
& 0.3336 & 0.0873 & 0.1906 & 0.8449
& 15.15 & 12.50 & 14.99 & 16.60
& 0.8454 & 0.9045 & 0.5871 & 0.5466 \\

MobileLLM-Pro 0-Shot
& 0.2079 & 0.0426 & \textbf{0.2034} & \textbf{0.8470}
& 15.72 & 12.60 & 14.81 & 17.29
& 0.8124 & 0.8832 & 0.4658 & 0.3320 \\

MobileLLM-Pro 1-Shot
& 0.1855 & 0.0334 & 0.1701 & 0.8423
& 16.20 & 12.96 & 15.40 & 17.56
& 0.8352 & 0.8965 & 0.4966 & 0.2284 \\

MobileLLM-Pro 3-Shot
& 0.1981 & 0.0684 & 0.1711 & 0.8306
& 15.91 & 12.94 & 15.81 & 17.47
& 0.8024 & 0.8532 & 0.4917 & 0.1951 \\
\hline
\end{tabular}%
}

\begin{flushleft}
\footnotesize
Qwen refers to Qwen3.5-0.8B, and MobileLLM-Pro refers to the MobileLLM-Pro-1B model.
\end{flushleft}
\end{table}

Abstract-based prompting achieved higher overall scores than full-text prompting across the evaluated zero-shot, one-shot, and three-shot settings for both models (Qwen3.5-0.8B: $p=.002$, $r_{rb}=.80$; MobileLLM-Pro-1B: $p<.001$, $r_{rb}=.74$). Among the evaluated prompting configurations, three-shot abstract prompting achieved the highest overall mean score for both models.

Abstract-based prompting was also computationally more efficient, being 8.28 times faster for Qwen3.5-0.8B and 8.00 times faster for MobileLLM-Pro-1B than full-text prompting (Table~\ref{tab:runtimes-hms}). These findings support H3 on PLOS in terms of overall summary quality and runtime.

\subsubsection*{H3: eLife}

Table~\ref{tab:few-shot-elife} presents the few-shot inference results for both models on eLife.

\begin{table}[!ht]
\centering
\caption{\textbf{Few-shot baseline results on the eLife test set.}}
\label{tab:few-shot-elife}
\resizebox{\textwidth}{!}{%
\begin{tabular}{@{}l*{12}{c}@{}}
\hline
& \multicolumn{4}{c}{\textbf{Relevance}}
& \multicolumn{4}{c}{\textbf{Readability}}
& \multicolumn{3}{c}{\textbf{Factuality}}
& \\
\cline{2-5}
\cline{6-9}
\cline{10-12}
\rotatebox{90}{\textbf{Setting}}
& \rotatebox{90}{\textbf{ROUGE\rotatebox{-90}{$\uparrow$}}}
& \rotatebox{90}{\textbf{BLEU\rotatebox{-90}{$\uparrow$}}}
& \rotatebox{90}{\textbf{METEOR\rotatebox{-90}{$\uparrow$}}}
& \rotatebox{90}{\textbf{BERTScore\rotatebox{-90}{$\uparrow$}}}
& \rotatebox{90}{\textbf{FKGL\rotatebox{-90}{$\downarrow$}}}
& \rotatebox{90}{\textbf{DCRS\rotatebox{-90}{$\downarrow$}}}
& \rotatebox{90}{\textbf{SMOG\rotatebox{-90}{$\downarrow$}}}
& \rotatebox{90}{\textbf{CLI\rotatebox{-90}{$\downarrow$}}}
& \rotatebox{90}{\textbf{SummaC\rotatebox{-90}{$\uparrow$}}}
& \rotatebox{90}{\textbf{FENICE\rotatebox{-90}{$\uparrow$}}}
& \rotatebox{90}{\textbf{LongDocFact\rotatebox{-90}{$\uparrow$}}}
& \rotatebox{90}{\textbf{Mean\rotatebox{-90}{$\uparrow$}}} \\
\hline
Qwen 3-Shot Abstract
& 0.3643 & 0.0605 & 0.1581 & \textbf{0.8373}
& \textbf{10.56} & \textbf{11.06} & 11.75 & \textbf{12.69}
& 0.7938 & 0.6995 & 0.5901 & \textbf{0.7441} \\

Qwen 1-Shot Abstract
& 0.3404 & 0.0658 & 0.1514 & 0.8372
& 10.87 & 11.50 & 12.72 & 13.60
& 0.7167 & 0.7364 & 0.6028 & 0.6639 \\

MobileLLM-Pro 3-Shot Abstract
& 0.2929 & 0.0632 & 0.1535 & 0.8297
& 11.90 & 11.83 & 13.04 & 15.65
& 0.9054 & 0.9145 & 0.6585 & 0.6589 \\

Qwen 0-Shot Abstract
& 0.3052 & 0.0623 & 0.1527 & 0.8359
& 10.68 & 11.18 & 11.51 & 12.80
& 0.5161 & 0.7093 & 0.5776 & 0.6319 \\

Qwen 1-Shot
& 0.3967 & 0.0786 & 0.1636 & 0.8173
& 12.10 & 12.00 & 14.28 & 14.99
& 0.8351 & \textbf{0.9472} & 0.4965 & 0.6230 \\

Qwen 0-Shot
& \textbf{0.3979} & \textbf{0.0820} & \textbf{0.1725} & 0.8220
& 11.94 & 12.13 & 12.45 & 12.92
& 0.6233 & 0.6556 & 0.4825 & 0.6033 \\

Qwen 3-Shot
& 0.3848 & 0.0742 & 0.1636 & 0.8147
& 12.90 & 12.68 & 14.56 & 14.48
& \textbf{0.9108} & 0.9325 & 0.4932 & 0.5721 \\

MobileLLM-Pro 1-Shot Abstract
& 0.2435 & 0.0547 & 0.1392 & 0.8294
& 12.10 & 13.17 & 14.30 & 17.25
& 0.9054 & 0.9143 & 0.6326 & 0.4497 \\

MobileLLM-Pro 0-Shot Abstract
& 0.2275 & 0.0513 & 0.1372 & 0.8308
& 14.97 & 12.62 & 15.22 & 16.58
& 0.8944 & 0.9327 & \textbf{0.7004} & 0.4427 \\

MobileLLM-Pro 0-Shot
& 0.2611 & 0.0519 & 0.1578 & 0.8203
& 15.01 & 12.04 & 13.34 & 15.94
& 0.7453 & 0.7252 & 0.5459 & 0.3873 \\

MobileLLM-Pro 1-Shot
& 0.2507 & 0.0422 & 0.1539 & 0.8125
& 17.53 & 12.22 & 12.97 & 16.89
& 0.6943 & 0.6852 & 0.5173 & 0.2366 \\

MobileLLM-Pro 3-Shot
& 0.2167 & 0.0435 & 0.1323 & 0.8230
& 15.64 & 12.86 & 15.18 & 17.46
& 0.5238 & 0.6744 & 0.5005 & 0.0954 \\
\hline
\end{tabular}%
}

\begin{flushleft}
\footnotesize
Qwen refers to Qwen3.5-0.8B, and MobileLLM-Pro refers to the MobileLLM-Pro-1B model.
\end{flushleft}
\end{table}

As observed on PLOS, abstract-based prompting achieved higher overall scores than full-text prompting across the evaluated zero-shot, one-shot, and three-shot settings for both models (Qwen3.5-0.8B: $p=.018$, $r_{rb}=.54$; MobileLLM-Pro-1B: $p<.001$, $r_{rb}=.782$). Three-shot abstract prompting achieved the highest overall mean score for both models.

The runtime advantage of abstract-based prompting was smaller on eLife than on PLOS. Nevertheless, abstract-based prompting was 2.49 times faster for Qwen3.5-0.8B and 2.48 times faster for MobileLLM-Pro-1B than full-text prompting (Table~\ref{tab:runtimes-hms}). These findings support H3 on eLife in terms of overall summary quality and runtime.

\section*{Discussion}

This study investigated the effects of input scope on biomedical lay summarization by comparing abstract-based and full-text-based approaches across two models and two datasets. The discussion is organized around the three hypotheses, focusing on the implications of the findings for summary quality and computational efficiency.

\subsection*{H1: Abstract-based vs. full-text LoRA fine-tuning}

Abstract-based LoRA fine-tuning achieved higher overall summary quality scores than full-text fine-tuning across both datasets and models, while reducing fine-tuning runtime by an average factor of 4.7. These findings suggest that using abstracts as fine-tuning input can offer a favorable balance between summary quality and computational cost in the evaluated settings.

This result differs from Basu et al. \cite{basu2025suwmit}, who reported better performance with full-text fine-tuning on several metrics using the eLife validation set. Their study also employed the larger Llama 3.1 Instruct 8B model. Differences in model architecture, model size, and experimental configuration may contribute to the contrasting findings, although the available evidence does not establish which factor accounts for the difference.

One possible explanation for the advantage of abstract-based fine-tuning is that abstracts provide a concise representation of the article's principal findings and conclusions. Their shorter length also reduces the amount of input processed during training. Full-text articles may contain additional details that are useful for generating comprehensive summaries, but they also require substantially more computation. The present results indicate that, under the evaluated conditions, this additional input did not translate into higher overall scores. Thus, abstract-based fine-tuning represents a computationally efficient alternative to full-text fine-tuning without sacrificing overall performance in these experiments.

\subsection*{H2: Three-shot abstract prompting vs. full-text LoRA fine-tuning}

Three-shot prompting with abstract--lay-summary examples achieved higher overall scores than full-text LoRA fine-tuning and was, on average, 8.58 times faster across the two datasets and models. This finding suggests that providing a small number of examples of the desired summarization style can be an effective alternative to fine-tuning on full-text articles.

The results differ from those of DeakinNLP \cite{to-etal-2024-deakinnlp}, where fine-tuning outperformed few-shot prompting. However, the methods were evaluated using different model families for the two approaches, limiting direct comparability. In the present study, three-shot prompting and LoRA fine-tuning were evaluated using the same models within a consistent experimental setup, providing a more controlled comparison of these approaches under the selected conditions.

A potential advantage of three-shot abstract prompting is that it demonstrates the desired lay-summary style through examples without requiring additional model training. Furthermore, using abstracts rather than full articles substantially reduces the amount of input processed during inference. Together, these characteristics may explain the observed computational advantage. The findings suggest that three-shot abstract prompting is a practical alternative to full-text LoRA fine-tuning when overall summary quality and runtime are considered jointly. However, the relative performance of prompting and fine-tuning may depend on the model, dataset, and evaluation criteria.

\subsection*{H3: Abstract vs. full-text few-shot prompting}

Abstract-based prompting achieved higher overall scores than full-text prompting across the evaluated zero-shot, one-shot, and three-shot settings, while being, on average, 5.31 times faster across both datasets and models. These results suggest that restricting the input to abstracts can improve computational efficiency while maintaining or improving overall summary quality in the evaluated prompting configurations.

Previous work, including 5cNLP \cite{lossio-ventura-etal-2025-5cnlp}, has investigated zero-shot and few-shot prompting with different demonstration settings. However, it did not examine the same matched comparison between abstract-based and full-text-based input. The present study therefore provides complementary evidence on how input scope affects prompting-based biomedical lay summarization.

The observed advantage of abstract-based prompting may be related to the concise nature of abstracts, which typically foreground the principal objectives, findings, and conclusions of a study. In contrast, full-text articles contain extensive methodological details and additional information that may not be necessary for producing a concise lay summary. Processing the full article also increases the input length and associated inference cost. Nevertheless, the current results do not establish whether the performance differences arise from reduced input length, differences in information content, or both. Further investigation would be needed to disentangle these factors.

Overall, the findings across H1--H3 indicate that abstract-based approaches achieved higher overall scores and lower computational costs than their full-text counterparts under the experimental conditions examined. This pattern was observed for both fine-tuning and prompting, suggesting that the advantages of abstract-based input are not limited to a single adaptation strategy. However, these conclusions should be interpreted within the scope of the two datasets and two models evaluated in this study.

\section*{Conclusion}

This study aimed to address a research gap by systematically comparing abstract-based fine-tuning and few-shot prompting with their full-text-based counterparts under a consistent experimental setup. The primary contribution is the finding that abstract-based approaches can achieve better overall lay summarization performance while substantially reducing runtime.

All three hypotheses were supported across both datasets and models. Abstract-based LoRA fine-tuning achieved higher overall lay summarization scores and was, on average, 4.70 times faster than full-text-based LoRA fine-tuning. Similarly, three-shot prompting using abstracts outperformed full-text-based LoRA fine-tuning while providing an 8.58-fold speedup. Furthermore, abstract-based prompting consistently outperformed full-text-based prompting across zero-shot, one-shot, and three-shot settings, with an average speedup of 5.31-fold.

These findings suggest that abstract-based approaches can offer an effective and computationally efficient alternative to full-text-based summarization, particularly when abstracts adequately capture the key information presented in the corresponding articles. By reducing the amount of input text, these approaches can lower computational costs and accelerate summary generation without compromising overall lay summary quality under the experimental conditions examined in this study.

This study has several limitations. First, the experiments were restricted to relatively small language models because of the computational demands associated with evaluating numerous experimental configurations, limited GPU resources, and model context-length constraints. Consequently, the extent to which these findings generalize to larger language models remains to be investigated. Second, the evaluation relied primarily on automatic metrics, and no clinician-based assessment was conducted. Future work should therefore investigate larger models and incorporate expert evaluations to assess the factual accuracy, clinical relevance, completeness, and readability of the generated lay summaries. Such evaluations would help establish whether the observed efficiency and performance advantages of abstract-based approaches extend to broader model settings and practical biomedical communication scenarios.

\section*{Supporting information}

\paragraph*{S1 File.}
\label{S1_File}
\textbf{Supplementary material.}
A separate supplementary document is provided with this article. It contains the complete prompt formulations and prompt-selection results. Three articles were randomly selected as few-shot examples from eLife \cite{elife14022,elife69174,elife30094}, and three from PLOS \cite{plos0000669,plos1003184,plos1007423}.

\bibliography{references}

\end{document}